%% file: IROSDirectionalCoordinates.tex
\documentclass[letterpaper, 10 pt, conference]{ieeeconf}

\IEEEoverridecommandlockouts                              

\usepackage{graphicx} 
\usepackage{epsfig} 
\usepackage{dblfloatfix} 
\usepackage{courier}
\usepackage{amsmath} 
\usepackage{amssymb}  
\usepackage{xcolor}
\usepackage{amsmath} 
\usepackage{amssymb}
\usepackage{float}
\usepackage{algorithm}		
\usepackage{algorithmicx} 	
\usepackage{algpseudocode}  
\usepackage{todonotes}
\usepackage{multicol}
\usepackage{paralist}
\usepackage[backend=bibtex,bibstyle=ieee,citestyle=numeric-comp,doi=false,isbn=false,url=false,eprint=false]{biblatex}
\title{\LARGE \bf
Localization with Directional Coordinates
}

\author{Charles Champagne Cossette, Mohammed Shalaby,  David Saussi\'e, James Richard Forbes 
\thanks{*This work was supported by the FRQNT under grant 2018-PR-253646, the Canadian Foundation for Innovation JELF, the William Dawson Scholar Program, and the NSERC Discovery Grant Program.}
\thanks{C. C. Cossette, M. Shalaby, J.R. Forbes are with the Department of Mech. Engineering, McGill University. { \small charles.cossette@mail.mcgill.ca, mohammed.shalaby@mail.mcgill.ca, james.richard.forbes@mcgill.ca}}%
\thanks{D. Saussi\'e is with the Department of Electrical Engineering, Polytechnique Montr\'eal. 
        {\small d.saussie@polymtl.ca}}%
}

\input{Commands.tex}

\definecolor{darkgreen}{rgb}{0,0.5,0}

\begin{document}
\input{arxiv_cover_ieee.tex}

\maketitle
\thispagestyle{empty}
\pagestyle{empty}

\begin{abstract}
    A coordinate system is proposed that replaces the usual three-dimensional Cartesian $x,y,z$ position coordinates, for use in robotic localization applications. Range, azimuth, and elevation measurement models become greatly simplified, and, unlike spherical coordinates, the proposed coordinates do not suffer from the same kinematic singularities and angle wrap-around. When compared to Cartesian coordinates, the proposed coordinate system results in a significantly enhanced ability to represent the true distribution of robot positions, ultimately leading to large improvements in state estimation consistency.
\end{abstract}

\section{Introduction}
This paper considers the problem of estimating the position of a robot, relative to some landmark or reference point, using range (distance), azimuth, and elevation (RAE) measurements. This problem has many applications in robotics, such as indoor localization or multi-robot relative position estimation. RAE measurements can be generated from a variety of sensors, such as radar used to track aircraft and spacecraft, or point cloud measurements obtained from LIDAR \cite[Ch.~6.4.3]{Barfoot2019}. Recently, ultra-wideband (UWB) radio has also been used to provide angle of arrival measurements \cite{Dotlic2018}, which is essentially an azimuth measurement, in addition to the usual distance measurements obtained between two UWB transceivers. However, single-antenna UWB angle of arrival measurements can have high variance \cite{Ledergerber2019}.

\begin{figure}[t]
    \centering
    \includegraphics[width = \linewidth, clip = true, trim = {1cm 1.0cm 1cm 0.9cm}]{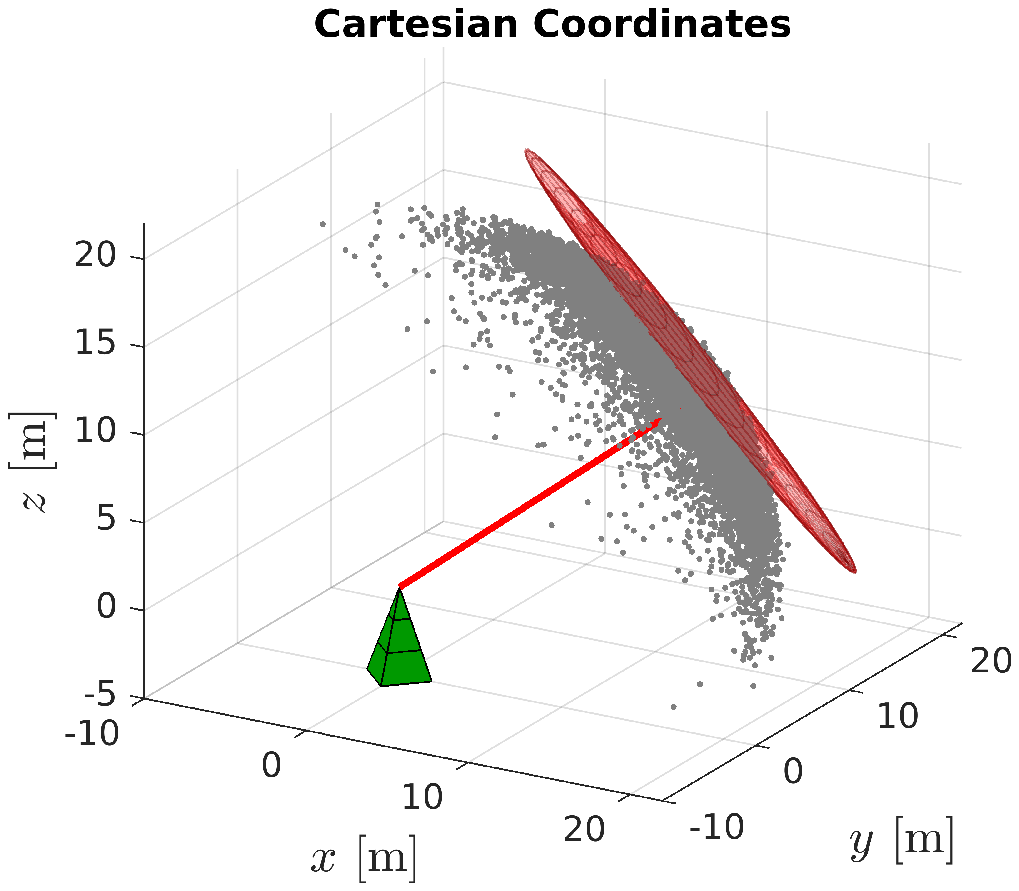}\\
    \includegraphics[width = \linewidth, clip = true, trim = {1cm 1.0cm 1cm 0.9cm}]{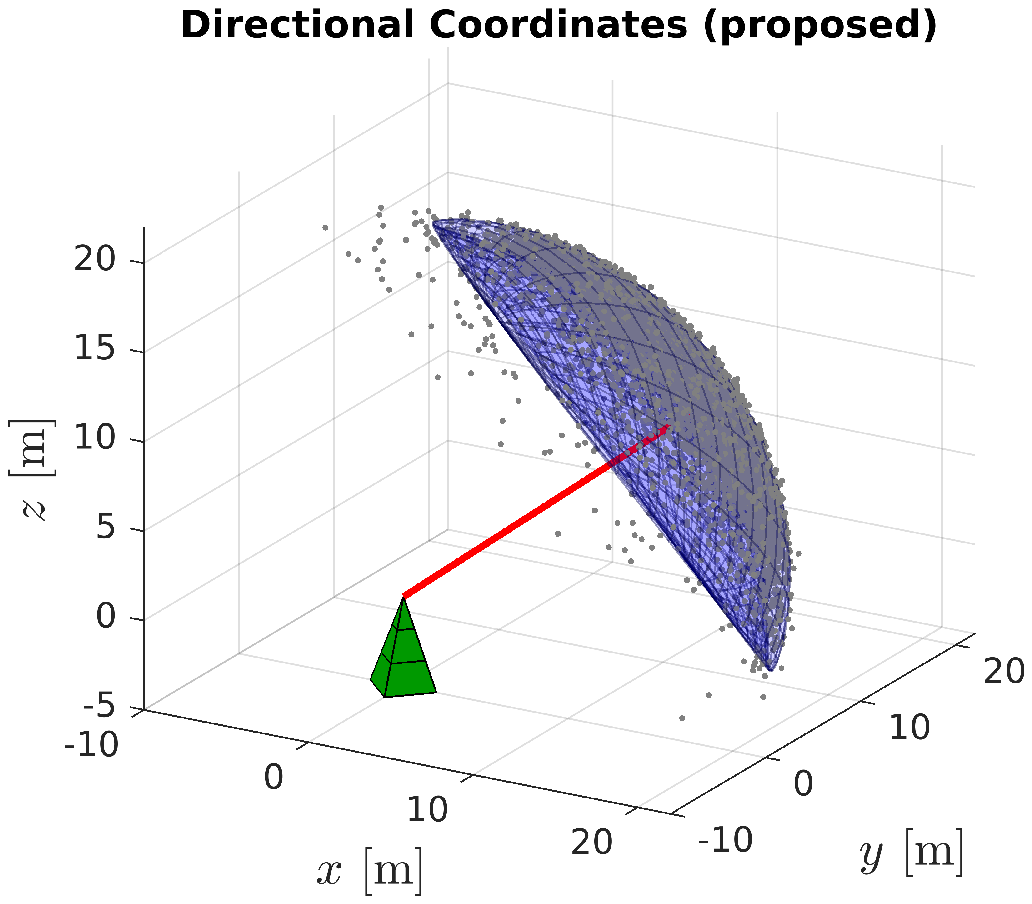}
    \vspace{-0.6cm}
    \caption{Distribution of positions of a robot given a single distance measurement (red line) to a landmark (green pyramid) and a Gaussian prior distribution. The point cloud represents the ``true'' distribution, as determined by a particle filter, while the red and blue volumes are $3\sigma$ equal probability contours in Cartesian and directional coordinates, respectively. }
    \label{fig:sphere_distribution}
\end{figure}

A typical probabilistic estimation algorithm will use the familiar Cartesian $x,y,z$ coordinates, with Gaussian estimators reporting the mean and covariance of the estimated distribution. However, as will be shown in this paper, Cartesian coordinates are generally a poor way of parameterizing the position of a robot when relying heavily on RAE measurements, especially when these measurements are very noisy, or if the robot position is highly uncertain. For example, the distribution of positions given a range measurement generally appears spherical, as shown in Figure~\ref{fig:sphere_distribution}, which  greatly differs from a standard Gaussian ellipsoid. For this reason, this paper proposes an alternate coordinate system, referred to herein as \emph{directional coordinates}. These coordinates are shown to provide a much more accurate representation of the position distribution, ultimately leading to improved state estimate consistency.

The idea of changing the coordinate system to better represent a distribution is not new. The ``banana distribution'' associated with robot position uncertainty, primarily resulting from attitude uncertainty and sensor noise, has motivated the use of \emph{exponential coordinates} \cite{Long2013}. A Gaussian distribution in exponential coordinates transforms to a ``banana'' distribution when mapped back to Cartesian coordinates. These exponential coordinates are naturally exploited when using matrix Lie groups in robotic estimation problems \cite{Barrau2017a,Barfoot2019}. Spherical coordinates use range, azimuth, and elevation as the coordinates themselves, and have been used in state estimation \cite{Boberg2009, Mallick2011}. Spherical coordinates are often modified to avoid singularities when the range is zero, such as using the logarithm of the range in \cite{Mallick2011}, or its reciprocal \cite{Boberg2009}. However, spherical coordinates possess an additional kinematic singularity at an elevation of $\varepsilon =\pm \pi/2\;\mathrm{rad}$, and also suffer from angle wrap-around, which must be overcome with explicit checks in the estimator implementation. Another strategy specific to RAE measurements is simply to directly convert them to a position measurement \cite{Lerro1993,Bordonaro2015}, thus creating a linear measurement model. However, this requires careful conversion of the original noise statistics to equivalent noise statistics on the new measurement, which can introduce inaccuracies.

The new \emph{directional coordinate} system proposed in this paper uses a range and direction cosine matrix (DCM) to parameterize position. The result is a smooth parameterization with 
\begin{inparaenum}
    \item no angle wrap-around issues,
    \item no kinematic singularities associated with the elevation,
    \item RAE measurement models that are linear, or have constant, state-estimate-independent Jacobians, and
    \item a filter that is significantly more consistent than a standard Cartesian coordinate filter, when noise levels are high.
\end{inparaenum} 
One drawback of the proposed use of directional coordinates is that a previously linear process model in Cartesian coordinates becomes nonlinear, an example being a simple ``velocity input'' process model. However, in applications with measurements arriving at a regular frequency, this added complexity is justified by the simplified measurement models, and enhanced ability to capture the true distributions. The proposed directional coordinates are evaluated in an extended Kalman filter (EKF) framework, and compared to an EKF that uses the usual Cartesian coordinates.  

The paper is as follows. Directional coordinates are defined in Section \ref{sec:dc}, with a basic probabilistic treatment shown in Section \ref{sec:prob}. Section \ref{sec:kinematics} considers directional coordinate kinematics. Simulation and experimental results are shown in Sections \ref{sec:sim} and \ref{sec:exp}, respectively.
\subsection{Notation}
A bolded $\mbf{1}$ or $\mbf{0}$ indicates an appropriately-sized identity  or zero matrix, respectively. A Gaussian distribution with mean $\mbs{\mu}$ and covariance $\mbs{\Sigma}$ is written as $\mc{N}(\mbs{\mu}, \mbs{\Sigma})$. The standard Euclidean 2-norm is denoted $\norm{\cdot}$. The \emph{special orthogonal group} of dimension 3 is denoted $SO(3)$ and whose elements belong to the set $\{ \mbf{C} \in \mathbb{R}^{3 \times 3} \;| \; \mbf{C}^\trans \mbf{C} = \mbf{1}, \; \det \mbf{C} = 1\}$. The operator $(\cdot)^\times:\mathbb{R}^3 \to \mathfrak{so}(3)$ is the skew-symmetric cross-product matrix operator, with $\mathfrak{so}(3)$ being the Lie algebra of $SO(3)$. To reduce clutter, the argument of time $(t)$ will be omitted from time-varying objects. 
\section{Directional Coordinates} \label{sec:dc}
In this paper, \emph{directional coordinates} are defined as the pair $(\rho, \mbf{C})$, where $\rho \in \mathbb{R}_{\geq0}$ and $\mbf{C} \in SO(3)$ is a direction cosine matrix (rotation matrix). The idea is to use these coordinates as an alternate representation of the components of a Cartesian position vector, denoted $\mbf{r} \in \mathbb{R}^3$. This is done with the defining relation
\beq \label{eq:dc}
\mbf{r} = \rho \mbf{C} \mbf{e}_1,
\eeq 
where $\mbf{e}_1 = [1\;0\;0]^\trans$. The \emph{range} $\rho = \norm{\mbf{r}}$ is the length of the position vector, whereas $\mbf{C}$ can be interpreted as a direction cosine matrix (DCM) that ``rotates'' the vector $\mbf{e}_1$ to produce a \emph{unit direction vector} $\mbs{\varrho} = \mbf{C} \mbf{e}_1 = \mbf{r} / \norm{\mbf{r}}$ collinear to $\mbf{r}$.

Directional coordinates are similar to spherical coordinates, but the azimuth and elevation angles are instead represented with a DCM $\mbf{C}$. This avoids the kinematic singularities associated with spherical coordinates, as well as any angle wrap-around issues.

\subsection{Cartesian to directional coordinates} \label{sec:cart_to_dc}
Given a position vector $\mbf{r} = [r_x \; r_y\; r_z]^\trans$ expressed in Cartesian coordinates, equivalent directional coordinates $(\rho, \mbf{C})$ satisfying \eqref{eq:dc} can be obtained from the well-known axis-angle parameterization of a DCM. The range is straightforwardly obtained with $\rho = \norm{\mbf{r}}_2$, while the direction cosine matrix $\mbf{C}$ can be obtained by first defining $\mbf{a}$ and $\psi$ as
\begin{align*}
\mbf{a} &= \frac{\rho \mbf{e}_1^\times \mbf{r}}{\norm{\rho \mbf{e}_1^\times \mbf{r}}} =  \bma{c}0 \\-r_z \\ r_y \ema \bigg/ \norm{ \bma{c}0 \\-r_z \\ r_y \ema},\\
\psi &= \arccos\frac{\rho \mbf{e}_1^\trans \mbf{r}}{\norm{\rho\mbf{e}_1} \norm{\mbf{r}}} = \arccos\frac{\rho \mbf{e}_1^\trans \mbf{r}}{\rho^2} = \arccos \frac{r_x}{\rho},
\end{align*}
from which $\mbf{C}$ is then given by $\mbf{C} = \exp(\psi \mbf{a}^{\times}),$ which is well defined except when $r_z = r_y = 0$. In this case, setting $\rho = \norm{\mbf{r}}$ and $\mbf{C} = \mbf{1}$ will satisfy the definition given by \eqref{eq:dc}. As such, a continuous function $\mbf{h}:\mathbb{R}^3 \to \mathbb{R} \times SO(3)$ is defined that performs the above operations to go from $\mbf{r}$ to $(\rho, \mbf{C})$. That is,
\beq \label{eq:cart_to_dc}
(\rho, \mbf{C}) = \mbf{h}(\mbf{r}).
\eeq
\subsection{Local parameterization} \label{sec:local_param}
Estimation tools that operate on $SO(3)$ directly are very well established \cite{Barfoot2019,Long2013,Barrau2017a}, most of which require the use of a \emph{local parameterization} of $SO(3)$. In this paper, this is given by $\mbs{\phi} = [\phi_1\; \phi_2]^\trans \in \mathbb{R}^2$, defined such that 
\bdis
\mbf{C} = \exp(\mbs{\phi}^\wedge),
\edis
where the \emph{wedge} operator $(\cdot)^\wedge: \mathbb{R}^2 \to \mathfrak{so}(3)$ is given by 
\bdis
\bma{c} \phi_1 \\ \phi_2 \ema^\wedge= \bma{ccc} 0 & -\phi_2 & \phi_1 \\ \phi_2 & 0 & 0 \\ - \phi_1 & 0 & 0 \ema,
\edis
and otherwise expressed as $ \mbs{\phi}^\wedge = ([0 \; \mbs{\phi}^\trans]^\trans)^\times$. An inverse mapping, named the \emph{vee} operator, $(\cdot)^\vee: \mathfrak{so}(3) \to \mathbb{R}^2$ can be defined such that $((\mbs{\phi})^\wedge)^\vee = \mbs{\phi}$.
It is also useful to define an operator $(\cdot)^\odot:\mathbb{R}^3 \to \mathbb{R}^{3 \times 2}$ such that the identity $\mbs{\phi}^\wedge \mbf{a} = \mbf{a}^\odot \mbs{\phi}$ holds. With the above definition of the wedge operator, it can be shown that the $(\cdot)^\odot$ operator must be
\beq
\mbf{a}^\odot = \bma{c} a_1 \\ a_2 \\ a_3 \ema^{\odot} = \bma{cc} a_3 & -a_2 \\ 0 & a_1 \\ -a_1 & 0 \ema.
\eeq
A few useful identities can be derived, for $\mbf{b}, \mbf{c} \in \mathbb{R}^2$, by simple expansion into components,
\beq
    \mbf{b}^{\wedge^\trans} = - \mbf{b}^\wedge, \quad\;    (\mbf{b} + \mbf{c})^\wedge = \mbf{b}^\wedge + \mbf{c}^\wedge, \quad\;    (\mbf{b}^\wedge \mbf{c}^\wedge)^\vee = \mbf{0}. \label{eq:iden}
\eeq

\section{Distributions in Directional Coordinates} \label{sec:prob}
Consider the ``mean" or ``nominal" directional coordinates $(\check{\rho} , \mbfcheck{C})$ and the perturbations $\delta \rho$ and $\delta \mbs{\phi}$ such that 
\begin{align}
    \rho &= \check{\rho} + \delta \rho, \label{eq:prob1} \\
    \mbf{C} &= \mbfcheck{C} \exp(\delta \mbs{\phi}^\wedge). \label{eq:prob2}
\end{align}
Define
\beq 
\delta \mbf{x} = \bma{c}\delta \rho \\ \delta \mbs{\phi} \ema \in \mathbb{R}^3.
\eeq
 If $\delta \mbf{x}$ is a random variable drawn from $\delta \mbf{x}\sim \mc{N}(\mbf{0}, \mbfcheck{P})$, this will induce a distribution over the values of $(\rho, \mbf{C})$.

\begin{figure}[b]
    \vspace{-0.2cm}
    \includegraphics[width = \linewidth, clip = true, trim= {1.8cm 2cm 2.4cm 2.8cm}]{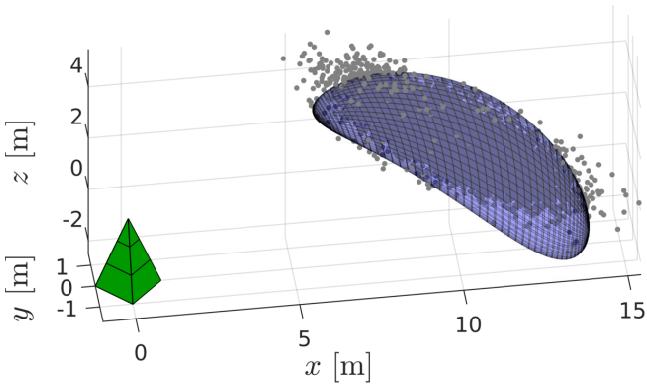}\\
    \vspace{-0.6cm}
    \caption{Samples from a Cartesian Gaussian distribution, shown as the point cloud. The bean-shaped volume shows a 3 standard deviation equal probability contour of a Gaussian distribution in directional coordinates, but mapped back to Cartesian coordinates for visualization. }
    \label{fig:prior}
\end{figure}

\subsection{Converting a Cartesian Gaussian distribution to directional coordinates} \label{sec:convert_prior}
A common requirement will be to convert a Gaussian distribution over Cartesian coordinates $\mbf{r} \sim \mc{N}(\mbfcheck{r}, \mbfcheck{P}_r)$, into an equivalent, or approximate, distribution in directional coordinates, represented with nominal point $(\check{\rho}, \mbfcheck{C})$ and corresponding distribution $\delta \mbf{x}\sim \mc{N}(\mbf{0}, \mbfcheck{P})$. Fortunately, a mapping $\mbf{h}(\cdot)$ has been identified in Section \ref{sec:cart_to_dc}, which allows for many well-known methods of passing a Gaussian through a nonlinearity \cite[Ch.~2.2.8]{Barfoot2019}. This paper uses sigma points for this step exclusively. 

A variety of sigma point methods, such as the \emph{unscented transform}, \emph{spherical cubature}, or \emph{Gauss-Hermite cubature} \cite[Ch.~6]{Sarkka2010}, can be used to generate a set of sigma points and corresponding weights $(\mbf{s}_i, w_i), \; i = 1,\ldots,N$ drawn from the prior distribution $\mc{N}(\mbfcheck{r}, \mbfcheck{P}_r)$. The Cartesian sigma points are transformed to directional coordinates through $\mbf{h}(\cdot)$,
\bdis
(\rho_i, \mbf{C}_i) = \mbf{h}(\mbf{s}_i), \qquad  i = 1,\ldots,N.
\edis
The covariance associated with the directional coordinates can then be approximated as
\bdis
\mbfcheck{P} \approx \sum_{i =1}^N w_i\delta\mbs{\xi}_i \delta\mbs{\xi}_i^\trans, \qquad \delta\mbs{\xi}_i = \bma{c} \rho_i - \check{\rho} \\ \ln(\mbfcheck{C}^\trans \mbf{C}_i)^\vee \ema.
\edis
In standard sigma point methods, $(\check{\rho}, \mbfcheck{C})$ would be obtained from the weighted mean of the transformed sigma point values $(\rho_i, \mbf{C}_i)$ \cite[Ch.~6]{Sarkka2010}, which would require an optimization procedure to find the mean $\mbfcheck{C}$ on $SO(3)$ \cite{Dai2010b}. However, as suggested by \cite{Brossard2020a}, and verified extensively through simulation in this paper, it is sufficient to simply obtain the nominal points $(\check{\rho},\mbfcheck{C})$ by passing $\mbfcheck{r}$ through the nonlinear model \eqref{eq:cart_to_dc}, $(\check{\rho},\mbfcheck{C}) = \mbf{h}(\mbfcheck{r})$,  without an apparent loss in accuracy. An example of the results of this sigma point procedure can be visualized in Figure \ref{fig:prior}.

\begin{figure}[b]
    
    \includegraphics[width = \linewidth, clip = true, trim = {0.3cm 0.5cm 1.3cm 0.49cm}]{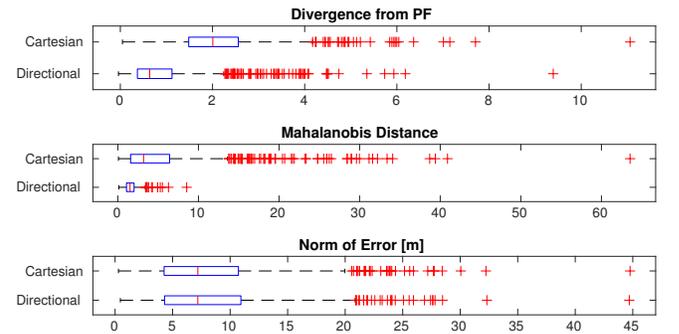}\\
    \vspace{-0.6cm}
    \caption{1000 Monte Carlo trials of a single Kalman correction given a distance measurement. Although the estimation accuracy using directional coordinates is not necessarily improved, directional coordinates offer a more consistent filter, as well as one that better resembles the particle filter.}
    \label{fig:single_step_mc}
\end{figure}
\subsection{Posterior distribution given a range measurement} \label{sec:post_range}
Consider now the task of estimating the distribution of $(\rho, \mbf{C})$ given a prior distribution and a single range measurement $y$ from the origin. The prior distribution's nominal point is denoted $(\check{\rho}, \mbfcheck{C})$, and is distributed as per \eqref{eq:prob1}-\eqref{eq:prob2} with $\delta \mbf{x} \sim \mc{N}(\mbf{0}, \mbfcheck{P})$. In directional coordinates, the range measurement model is linear,
\bdis
y = \rho + \nu, \qquad \nu \sim \mc{N}(0, R).
\edis
This leads to a trivial linearization procedure, where small changes in $\delta y$ are related to small $\delta \mbf{x}$ through $\delta y = \mbf{H} \delta \mbf{x} + \mbf{M}\nu$, with $\mbf{H} = [1 \; 0 \; 0]$ and $\mbf{M} = 1$. This provides a setup to use a \emph{multiplicative extended Kalman filter} \cite{Barfoot2019} correction step to estimate the posterior mean $(\hat{\rho},\mbfhat{C})$ and covariance $\mbfhat{P}$ with 
\begin{align}
    \mbf{K} &= \mbfcheck{P}  \mbf{H}^\trans(\mbf{H}\mbfcheck{P}\mbf{H}^\trans + \mbf{M}R\mbf{M}^\trans) ^{-1}, \label{eq:kalman1}\\
    \delta \mbfcheck{x} &= \bma{c} \delta \check{\rho} ,\\ \delta \mbscheck{\phi} \ema = \mbf{K}\mbf{z}, \qquad \mbf{z} = y - \check{\rho}, \\
    \hat{\rho} &= \check{\rho} + \delta \check{\rho}, \\ 
    \hat{\mbf{C}} &= {\mbfcheck{C}} \exp (\delta \mbscheck{\phi}^\wedge) ,\\
    \mbfhat{P} &= (\mbf{1} - \mbf{K} \mbf{H}) \mbfcheck{P}(\mbf{1} - \mbf{K} \mbf{H})^\trans + \mbf{K} \mbf{M} R \mbf{M}^\trans \mbf{K}^\trans\label{eq:kalman2}.
\end{align}
Figure \ref{fig:sphere_distribution} visualizes the results of this single EKF correction step when it is performed in Cartesian coordinates and directional coordinates, for the same Cartesian Gaussian prior. The point cloud is the distribution determined by a standard bootstrap particle filter \cite[Ch.~7]{Sarkka2010}, and is considered the closest approximation to the true distribution. In Cartesian coordinates, the measurement model is given by $y = \norm{\mbf{r}}_2 + \nu$, which is nonlinear, and requires linearization to produce a state-dependent measurement Jacobian. Linearization errors aside, Figure \ref{fig:sphere_distribution} clearly shows the degree to which the distribution is non-Gaussian, when expressed in Cartesian coordinates. As such, under high uncertainty, there is no Cartesian Gaussian estimator that could ever accurately represent this distribution.

This simple single-step correction experiment is repeated for 1000 Monte Carlo trials, where the prior mean $\mbfcheck{r}$ and covariance $\mbfcheck{P}_r$ are randomized, and the true position $\mbf{r}^{\mathrm{true}}$ is randomly sampled from this prior distribution. The prior distribution is converted to directional coordinates using Section \ref{sec:convert_prior}, and the posterior is calculated using Section \ref{sec:post_range}. Figure \ref{fig:single_step_mc} displays various metrics of the 1000 Monte Carlo trials. The ``dissimilarity'', or divergence, from the particle filter is calculated using the method in \cite{Wang2009}, which calculates an approximate Kullback-Leibler (KL) divergence between two point clouds, created by sampling their respective distributions. The Mahalanobis distance for the directional coordinates is calculated with 
\bdis
D = \sqrt{\delta\mbs{\xi}^\trans\mbfhat{P}^{-1} \delta\mbs{\xi}}, \qquad\delta\mbs{\xi} = \bma{c} \rho^{\mathrm{true}}-\hat{\rho}  \\ \ln(\mbfhat{C}^\trans \mbf{C}^{\mathrm{true}})^\vee \ema.
\edis
Although, on a single step, there is no accuracy improvement when using directional coordinates, there are substantially enhanced consistency properties, as reflected by the significantly lower Mahalanobis distances and divergence from the particle filter.

\subsection{Posterior distribution given azimuth and elevation measurements}  \label{sec:post_ae}
Another set of measurements that are naturally well-suited to directional coordinates are azimuth and elevation (AE) measurements, denoted $\alpha$ and $\varepsilon$ respectively. Azimuth and elevation measurements are defined to produce a direction vector $\mbs{\varrho}$ where
\beq \label{eq:direction_vector}
\mbs{\varrho} =\mbf{C}_z(\alpha)^\trans \mbf{C}_y(-\varepsilon)^\trans \mbf{e}_1, 
\eeq
and $\mbf{C}_z(\cdot),\; \mbf{C}_y(\cdot)$ are principle rotation DCMs about the $z$ and $y$ axes, respectively. Hence, if the direction vector resulting from \eqref{eq:direction_vector} is considered to be the measurement itself, the measurement model
\bdis
\mbf{y} = \mbf{C} \mbf{e}_1 + \mbs{\nu}, \qquad \mbs{\nu} \sim \mc{N}(\mbf{0}, \mbf{R}).
\edis
is obtained. The covariance $\mbf{R}$ can be obtained from covariances associated with $\alpha$ and  $\varepsilon$ using a standard linearization procedure \cite[Ch.~2.2.8]{Barfoot2019}. Next, a measurement innovation $\mbf{z}$ is defined following inspiration from the \emph{invariant extended Kalman filter} (IEKF) literature \cite{Barrau2017a}. Moreover, only two components of $\mbf{z}$ are used to avoid creating a fictitious third measurement, when only two are actually measured, that being $\alpha$ and $\varepsilon$. This is done through use of the projection matrix $\mbf{E}$ in \eqref{eq:innovation}, where the innovation is defined as
\begin{align}
    \mbf{z} &\triangleq \underbrace{\bma{ccc} 0 & 1 & 0 \\ 0 & 0 & 1 \ema}_\mbf{E} \mbfcheck{C}^\trans(\mbf{y} - \mbfcheck{y}) \label{eq:innovation}\\
    &=  \mbf{E} \mbfcheck{C}^\trans\mbf{C} \mbf{e}_1 + \mbf{E} \mbfcheck{C}^\trans\mbs{\nu} - \mbf{E} \mbfcheck{C}^\trans\mbfcheck{C} \mbf{e}_1  \nonumber\\
    &\approx\mbf{E} (\mbf{1} + \delta \mbs{\phi}^\wedge)\mbf{e}_1 + \mbf{E} \mbfcheck{C}^\trans\mbs{\nu}- \mbf{E}\mbf{e}_1 \nonumber\\
    &= \mbf{E} \mbf{e}_1^\odot \delta \mbs{\phi} + \mbf{E} \mbfcheck{C}^\trans \mbs{\nu} \triangleq \mbf{H}_\phi \delta \mbs{\phi} + \mbf{M} \mbs{\nu}, \label{eq:ae_innov}
\end{align}
where $\mbf{C}=\mbfcheck{C}\exp(\delta \mbs{\phi}^\wedge)$, and the first-order approximation $\exp(\delta \mbs{\phi}^\wedge) \approx (\mbf{1} + \delta \mbs{\phi}^\wedge)$ has been made. As with the range measurements, the measurement Jacobian $\mbf{H} = [\mbf{0}\; \mbf{H}_\phi]$ is constant and state-estimate independent, and would otherwise be highly nonlinear in Cartesian coordinates. One may return to \eqref{eq:kalman1}-\eqref{eq:kalman2} to compute a Kalman filter correction step, with new definitions for $\mbf{y}, \;\mbf{z},\; \mbf{H}, \;\mbf{M},$ and $\mbf{R}$.

\begin{figure*}[t]
    \includegraphics[width = 0.33\textwidth, clip = true, trim = {0.65cm 0cm 0.8cm 0.3cm}]{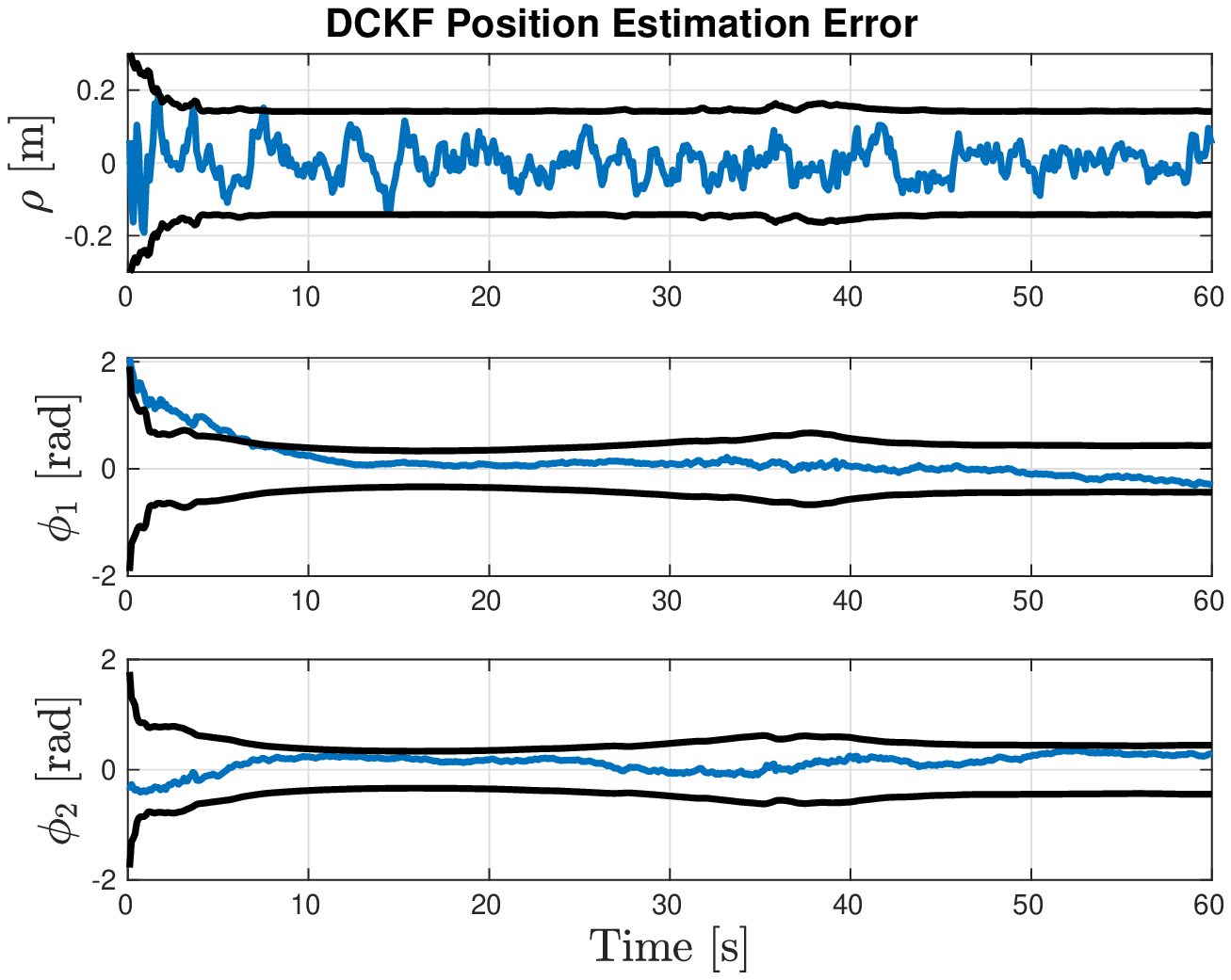}%
    \includegraphics[width = 0.33\textwidth, clip = true, trim = {0.65cm 0cm 0.8cm 0.3cm}]{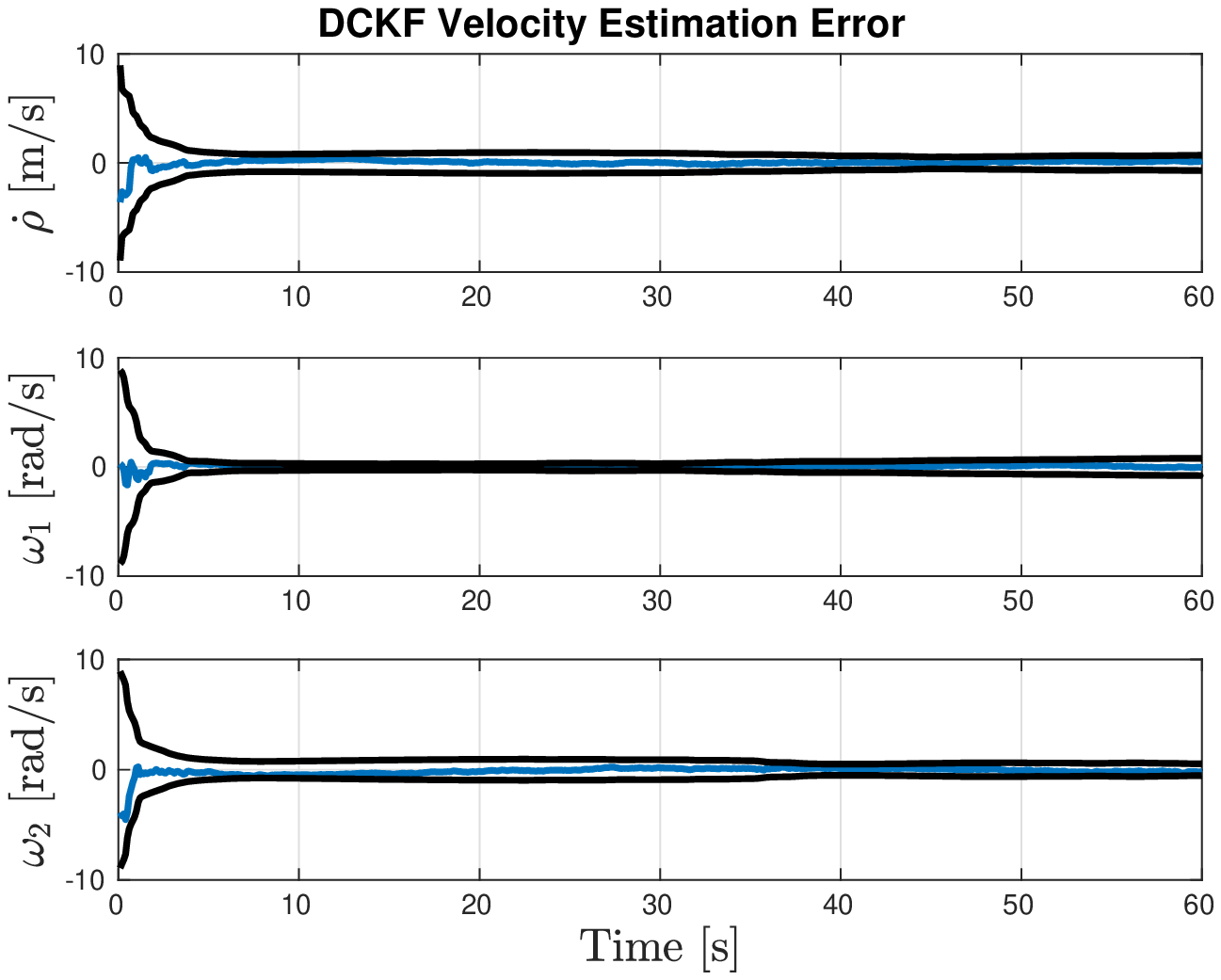}%
    \includegraphics[width = 0.33\textwidth, clip = true, trim = {0.65cm 0cm 0.8cm 0.3cm}]{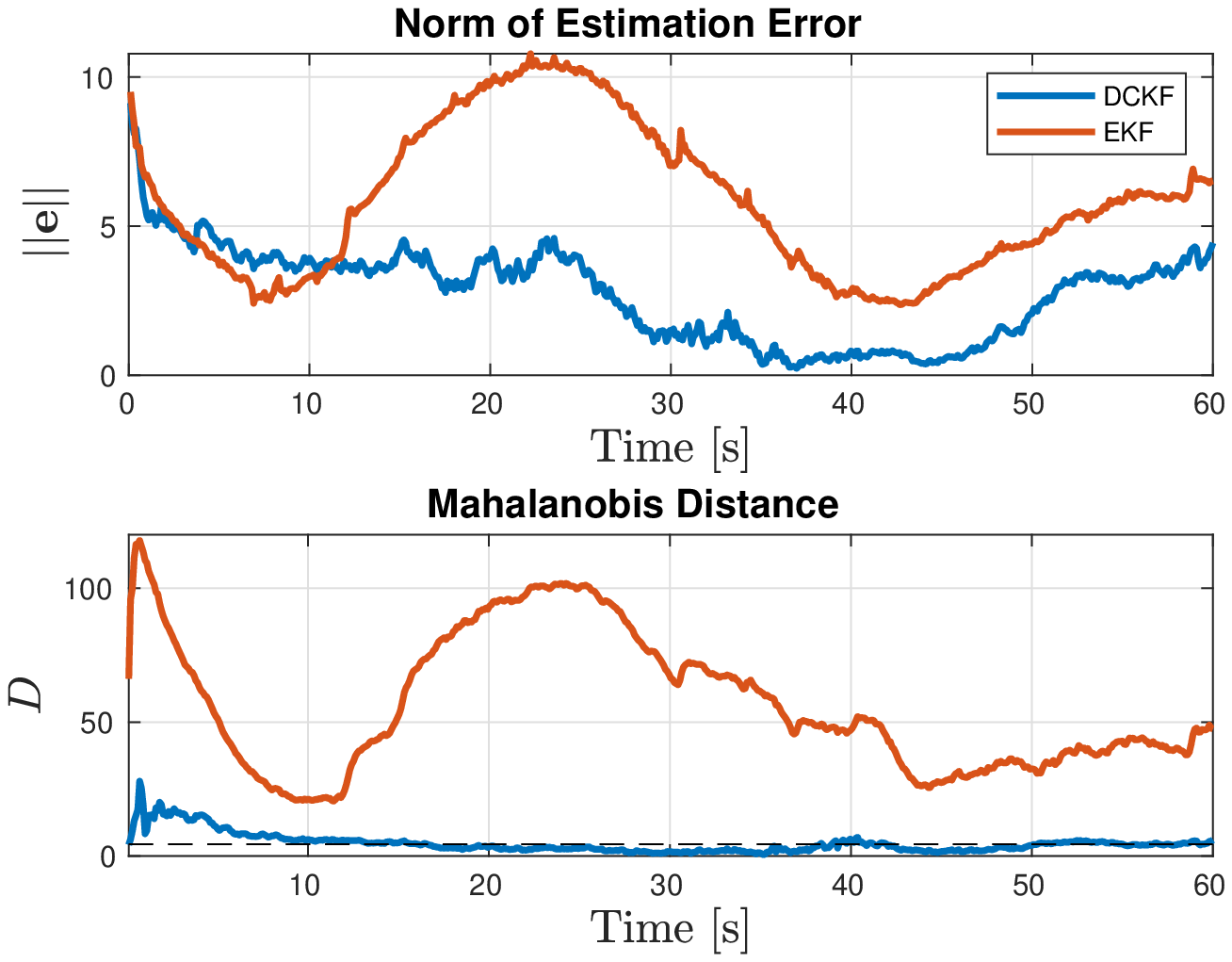}%
    \vspace{-0.4cm}
    \caption{Simulation results from a single trial. Black lines represent the $\pm3\sigma$ confidence bounds. The total estimation error on the right figure is calculated for both filters with $\mbf{e} = [\mbf{r}^{\mathrm{true}^\trans} \; \mbf{v}^{\mathrm{true}^\trans}]^\trans-[\mbfhat{r}^\trans\; \mbfhat{v}^\trans]^\trans$.}
    \label{fig:sim_trial}
\end{figure*}
\section{Directional Coordinate Kinematics} \label{sec:kinematics}
To execute dynamic filtering tasks, a process model is required, and it will usually be necessary to relate the rate of change of the directional coordinates $(\dot{\rho}, \mbfdot{C})$ to a Cartesian velocity input $\mbf{v} = \mbfdot{r}$. Differentiating \eqref{eq:dc} with respect to time,
\beq \label{eq:vel1}
\mbfdot{r} = \dot{\rho} \mbf{C} \mbf{e}_1 + \rho \mbfdot{C} \mbf{e}_1.
\eeq
The time rate of change of $\mbf{C}$ is $\mbfdot{C} = \mbf{C}\mbs{\omega}^\wedge$, where $\mbs{\omega} \in \mathbb{R}^2$. Substituting $\mbfdot{C} = \mbf{C}\mbs{\omega}^\wedge$ into \eqref{eq:vel1}, and using the $(\cdot)^\odot$ operator defined in Section \ref{sec:local_param} results in
\begin{align}
    \mbfdot{r} &= \dot{\rho} \mbf{C} \mbf{e}_1 + \rho \mbf{C}\mbs{\omega}^\wedge \mbf{e}_1  \nonumber\\
    &=  \dot{\rho} \mbf{C} \mbf{e}_1 + \rho \mbf{C}\mbf{e}_1^\odot \mbs{\omega}  \nonumber\\
    &= \bma{cc} \mbf{C} \mbf{e}_1 &\rho \mbf{C} \mbf{e}_1^\odot \ema \bma{c} \dot{\rho} \\ \mbs{\omega} \ema \nonumber\\
    &= \underbrace{ \mbf{C}   \bma{cc} \mbf{e}_1 &\mbf{e}_1^\odot \ema \bma{cc} 1 & 0  \\ 0 & \rho \mbf{1}\ema}_{\mbf{S}(\rho,\mbf{C})} \bma{c} \dot{\rho} \\ \mbs{\omega} \ema. \label{eq:vel2}
\end{align}
To obtain the time rate of change of the directional coordinates, $\mbf{S}$ must be invertible. Fortunately, assuming $\rho >0$, $\mbf{S}$ has an analytical inverse given by
\begin{align*}
    \mbf{S}(\rho,\mbf{C})^{-1} &= \bma{cc} 1 & 0  \\ 0 & (1/\rho) \mbf{1}\ema \bma{c} \mbf{e}_1^\trans \\ \mbf{e}_1^{\odot^\trans} \ema\mbf{C}^\trans.
\end{align*}
It follows that 
\beq \label{eq:vel3}
\bma{c} \dot{\rho} \\ \mbs{\omega} \ema = \mbf{S}(\rho,\mbf{C}) ^{-1} \mbf{v},
\eeq
which is well defined provided $\rho > 0$. Equation \eqref{eq:vel3} can be easily split into its two constituent equations, leading to the time rate of change of the directional coordinates, that being
\beq\label{eq:vel_rho}
    \dot{\rho} = \mbf{e}_1^\trans \mbf{C}^\trans \mbf{v}, \qquad    \mbfdot{C} = \frac{1}{\rho} \mbf{C}(\mbf{e}_1^{\odot^\trans} \mbf{C}^\trans \mbf{v})^\wedge. 
\eeq
The only kinematic singularity is located at $\rho = 0$, as opposed to spherical coordinates, which have an additional singularity at an elevation of $\varepsilon = \pm \pi/2\; \mathrm{rad}$. This singularity is not of concern provided the robot device does not move near the origin. This is often the case in practice, as if another robot or measurement device is located at the origin, the hardware itself forces a minimum separating distance.

\subsection{Linearization}
It is necessary to linearize the directional coordinate kinematics in order to execute an EKF prediction step. This is done by applying small perturbations to $\rho, \mbf{C}, \mbf{v}$ with $\rho = \hat{\rho} + \delta \rho$, $\mbf{C} \approx \mbfhat{C} (\mbf{1} + \delta \mbs{\phi}^\wedge)$, and $\mbf{v} = \mbfhat{v} + \delta \mbf{v}$. Equation \eqref{eq:vel_rho} becomes
\begin{align} \label{eq:lin3}
    \dot{\hat{\rho}} + \delta \dot{\rho} &\approx \mbf{e}_1^\trans \left( \mbfhat{C}(\mbf{1} + \delta \mbs{\phi}^\wedge)\right)^\trans (\mbfhat{v}+ \delta \mbf{v}).
\end{align}
Expanding \eqref{eq:lin3}, neglecting higher-order terms, and applying identities \eqref{eq:iden}, eventually gives
\bdis
\delta \dot{\rho}  \approx  - \mbf{e}_1^\trans  (\mbfhat{C}^\trans \mbfhat{v})^{\odot} \delta \mbs{\phi} + \mbf{e}_1^\trans \mbfhat{C}^\trans \delta \mbf{v} .
\edis
Linearizing the DCM kinematics gives 
\begin{align}
    \frac{\dee (\mbf{1} + \delta \mbs{\phi}^\wedge)}{\dee t} &\approx \frac{\dee(\mbfhat{C}^\trans \mbf{C})}{\dee t} = \dot{\mbfhat{C}}^\trans \mbf{C} + \mbfhat{C}^\trans \mbfdot{C} ,
\end{align}
\beq
    \delta \mbsdot{\phi}^\wedge \approx  \frac{-1}{\hat{\rho}} (\mbf{e}_1^{\odot^\trans} \mbfhat{C}^\trans \mbfhat{v})^\wedge \mbfhat{C}^\trans \mbfhat{C}(\mbf{1} + \delta \mbs{\phi}^\wedge) + \frac{1}{{\rho}} \mbfhat{C}^\trans \mbf{C}(\mbf{e}_1^{\odot^\trans} \mbf{C}^\trans \mbf{v})^\wedge. \label{eq:lin1}
\eeq
Again, after expansion, neglecting higher-order terms, applying identities \eqref{eq:iden}, and application of the $(\cdot)^\vee$ to both sides results in
\beq
\delta \mbsdot{\phi} \approx \frac{1}{\hat{\rho}} \mbf{e}_1^{\odot^\trans} \mbfhat{C}^\trans\delta \mbf{v} - \frac{1}{\hat{\rho}}\mbf{e}_1^{\odot^\trans} ( \mbfhat{C}^\trans \mbfhat{v})^\odot \delta  \mbs{\phi}  - \frac{1}{\hat{\rho}^2}  \mbf{e}_1^{\odot^\trans}\mbfhat{C}^\trans \mbfhat{v} \delta \rho. \label{eq:lin2}
\eeq
\begin{figure}[t]
    \centering
    \includegraphics[width = \linewidth, clip = true, trim = {0.7cm 0cm 0.8cm 0.48cm}]{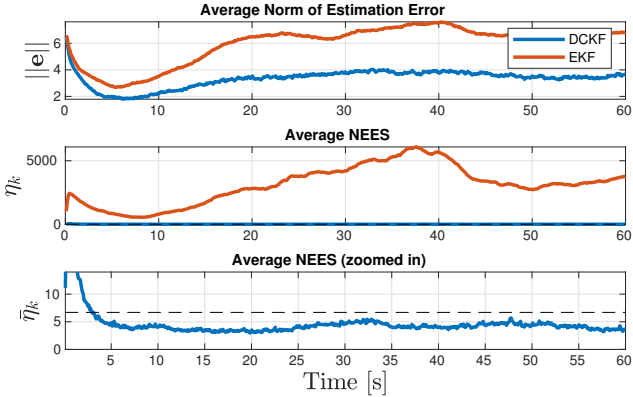}%
    \vspace{-0.4cm}
    \caption{Average estimation error and NEES, throughout time, for 100 Monte Carlo trials. The dashed line represents the one-sided 99.7\% probability boundary. }
    \label{fig:sim_mc}
\end{figure}
\section{Simulation} \label{sec:sim}
Two different EKFs are tested in simulation. The only difference between the two EKFs is the coordinate system used for the position parameterization. The standard EKF uses Cartesian coordinates, whereas the \emph{directional coordinate Kalman filter} (DCKF) uses directional coordinates. In these simulations, it is assumed that the acceleration $\mbf{a} = \mbfdot{v} = \ddot{\mbf{r}}$ of some point is measured with noise, along with noisy RAE measurements. The EKF state consists of $[\mbf{r}^\trans\; \mbf{v}^\trans]^\trans$, while the DCKF state is $(\rho, \mbf{C}, \mbf{v})$, where $\mbf{v} \in \mathbb{R}^3$ is a Cartesian velocity.

For the DCKF, the continuous-time linearized process model can be written as $\delta \mbfdot{x} = \mbf{A}\delta\mbf{x} + \mbf{L} \delta \mbf{a}$ where $\delta \mbf{x} = [\delta \rho \; \delta \mbs{\phi}^\trans \; \delta \mbf{v}^\trans]^\trans$, 
\beq
\mbf{A} = \bma{ccc} 0 & - \mbf{e}_1^\trans  (\mbfhat{C}^\trans \mbfhat{v})^{\odot} & \mbf{e}_1^\trans \mbfhat{C}^\trans \\  - \frac{1}{\bar{\rho}^2}  \mbf{e}_1^{\odot^\trans}\mbfhat{C}^\trans \mbfhat{v}  &  - \frac{1}{\bar{\rho}}\mbf{e}_1^{\odot^\trans} ( \mbfhat{C}^\trans \mbfhat{v})^\odot& \frac{1}{\bar{\rho}}  \mbf{e}_1^{\odot^\trans}\mbfhat{C}^\trans \\ \mbf{0} & \mbf{0} & \mbf{0}\ema, \label{eq:double_int_lin}
\eeq
and $\mbf{L} = [0 \; \mbf{0} \; \mbf{1}]^\trans$. The prediction step of the DCKF integrates \eqref{eq:vel_rho} and $\mbfdot{v} = \mbf{a}$ forward in time with simple Euler integration, but any method can be used. An equivalent discrete-time linearized model $\delta \mbf{x}_k = \mbf{A}_{k-1} \delta \mbf{x}_{k-1} + \mbf{w}_{k-1}, \;\mbf{w}_{k-1} \sim \mc{N}(\mbf{0}, \mbf{Q}_{k-1})$ can be obtained for the DCKF using a standard discretization technique such as a zero-order-hold. The matrices $\mbf{A}_{k-1}$ and $ \mbf{Q}_{k-1}$ are used to propagate the filter covariance forwards with the standard expression, $\mbfcheck{P}_k = \mbf{A}_{k-1} \mbfhat{P}_{k-1} \mbf{A}_{k-1}^\trans + \mbf{Q}_{k-1}$. The correction step for the DCKF is performed using equations \eqref{eq:kalman1}-\eqref{eq:kalman2}, with $\mbf{y}, \;\mbf{z},\; \mbf{H}, \;\mbf{M},$ and $\mbf{R}$ matrices defined in Sections \ref{sec:post_range} for a range measurement, or \ref{sec:post_ae} for azimuth-elevation measurements. The $\mbf{H}$ matrices are augmented with zeros due to the additional state $\mbf{v}$.

Figure \ref{fig:sim_trial} shows an example run with no covariance tuning, displaying the error behavior and consistency of the proposed DCKF. Figure \ref{fig:sim_mc} contains results for 100 Monte Carlo trials with mild covariance tuning. The \emph{normalized estimation error squared} (NEES) \cite[Ch.~5.4]{Bar-Shalom2001} $\eta_k^i$ is a consistency metric, calculated for the directional coordinates at time step $k$, on trial $i$, with
\beq
\eta_k^i= \delta\mbs{\xi}_k^\trans\mbfhat{P}^{-1}_k \delta\mbs{\xi}_k, \qquad\delta\mbs{\xi}_k = \bma{c} \rho^{\mathrm{true}}_k - \hat{\rho}_k \\ \ln(\mbfhat{C}_k^\trans \mbf{C}^{\mathrm{true}}_k)^\vee \\ \mbf{v}^\mathrm{true}_k - \mbfhat{v}_k \ema.
\eeq
 Under large noise and initial uncertainty, the DCKF boasts an average 44\% reduction in estimation error compared to the EKF and remains consistent, whereas the EKF is completely inconsistent, even with dedicated covariance tuning.

\begin{figure}[t]
\begin{minipage}{0.4\linewidth}
    \begin{figure}[H]
        \centering
        \includegraphics[width = 0.9\linewidth, clip = true, trim = {8cm 3cm 8cm 3cm}]{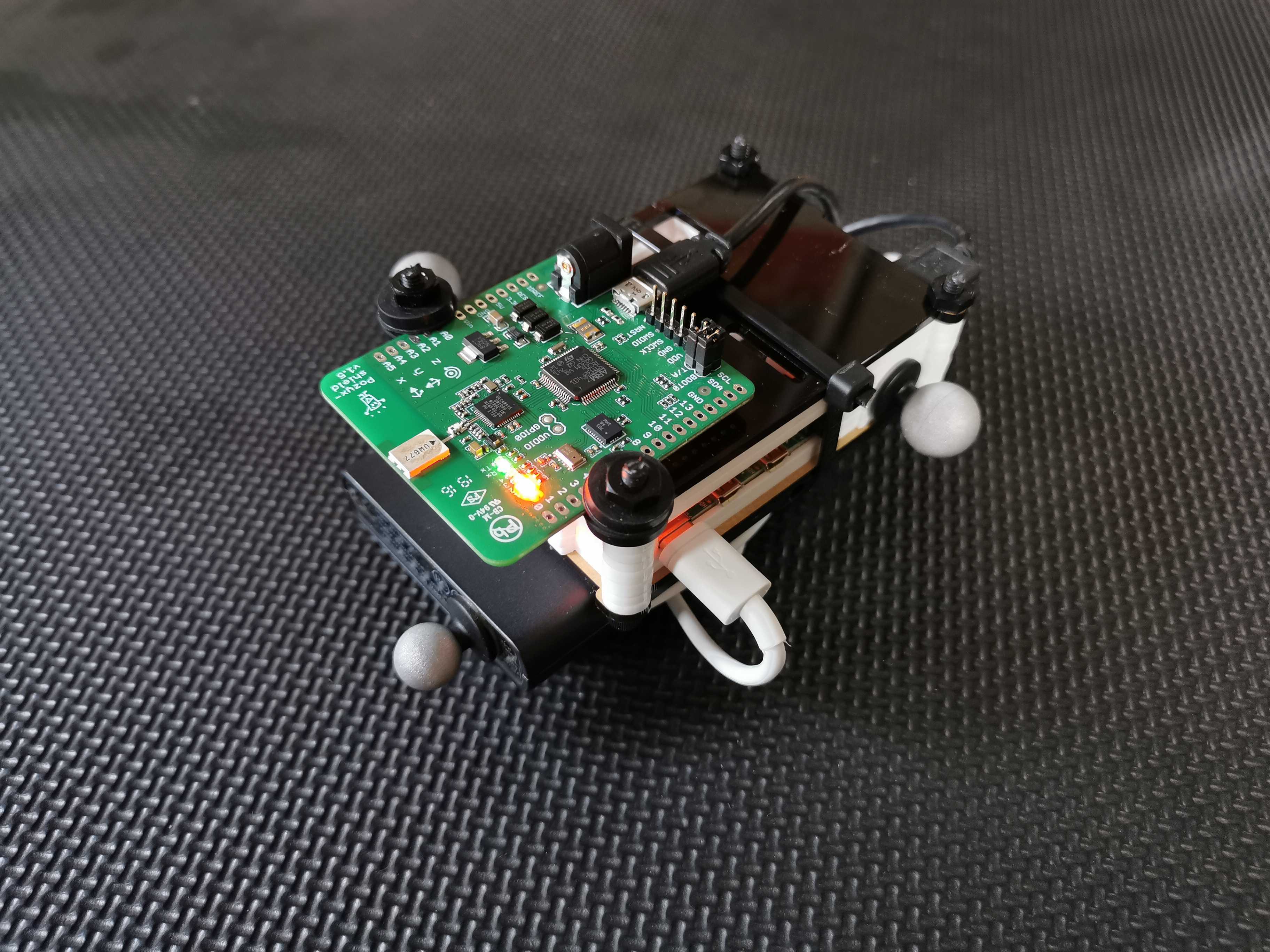}
        \caption{Pozyx UWB Developer Tag mounted to a Raspberry Pi 4B.}
        \label{fig:baby_agent}
    \end{figure}
\end{minipage}%
\begin{minipage}{0.6\linewidth}
\begin{table}[H]
    \caption{Simulation Noise Properties}
    \vspace{-0.2cm}
    \begin{center}
    \begin{tabular}{c|c|c}
    \hline
    \textbf{Specification} & \textbf{Value} & \textbf{Units}\\
    \hline \hline
    Range std. dev. & 0.1 & $ \mathrm{m}$ \\
    AE std. dev. & 0.8 & $\mathrm{rad}$ \\
    Accel. std. dev & 0.1 & $\mathrm{m/s}^2$\\
    Init. pos. std. dev. & $5$ & $\mathrm{m}$\\
    Init. vel. std. dev. & $3$ & $\mathrm{m/s}$\\
    Meas. frequency & $10$ & $\mathrm{Hz}
    $\\
    \hline
    \end{tabular}
    \label{table:sim1}
    \end{center}
    \vspace{-8pt}
\end{table}%
\end{minipage}
\vspace{-0.2cm}
\end{figure}
\begin{figure}[t]
    \centering
    \includegraphics[width = \linewidth,clip = true, trim = {0.45cm 0cm 1cm 0cm}]{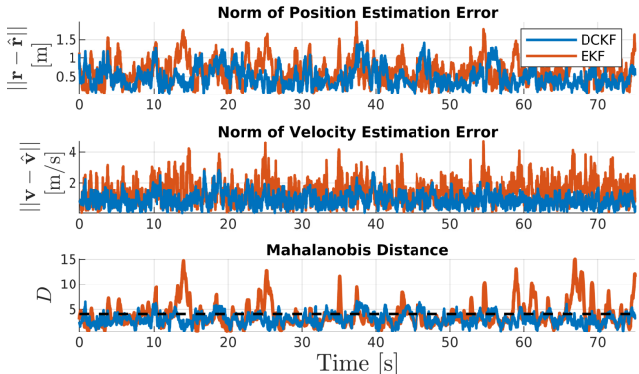}
    \vspace{-0.8cm}
    \caption{Experimental results from a single trial, with the dashed line representing the one-sided 99.7\% probability boundary.}
    \label{fig:exp_results}
\end{figure}

    \section{Experiment} \label{sec:exp}
The EKF and DCKF are also compared in a real experiment. Figure \ref{fig:baby_agent} shows the Raspberry Pi 4B that was used with an LSM9DS1 IMU, providing accelerometer and gyroscope measurements, and a Pozyx UWB Developer Tag, providing distance measurements to a Pozyx UWB Creator Anchor on the floor. Because the particular UWB modules used do not provide azimuth and elevation measurements, azimuth and elevation measurements were simulated using ground-truth data, with artificially added zero-mean Gaussian noise with a standard deviation of $0.8\;\mathrm{rad}$. This noise level is intentionally high, to induce a performance difference between the DCKF and EKF, and highlight the DCKF's improvement under large noise. The device moved in an overall volume of roughly $5~\mathrm{m} \times 4~\mathrm{m} \times 2~\mathrm{m}$. Ground truth position and attitude measurements are collected using an OptiTrack optical motion capture system.

Figure \ref{fig:exp_results} shows various performance and consistency metrics associated with the experimental trial. The position and velocity root-mean-squared error (RMSE) was ${0.55\;\mathrm{m}}$, ${0.87\;\mathrm{m/s}}$ for the DCKF, and ${0.97\;\mathrm{m}}$, ${1.56\;\mathrm{m/s}}$ for the EKF, respectively. This corresponds to a 43\% reduction in position RMSE, and a 44\% reduction in velocity RMSE. Once again, the DCKF is both more accurate and more consistent, when compared to the EKF.

\section{Conclusion}
This paper introduces a new coordinate system for parameterizing positions, and shows how it is particularly well-suited for RAE measurements. With these measurements, the new directional coordinate system captures the posterior distribution much more effectively than Cartesian coordinates, which culminates in a more accurate and significantly more consistent estimator. Future work could investigate eliminating the kinematic singularity at $\rho = 0$, by using the logarithm of the range as a coordinate, as done in \cite{Mallick2011}.

{\AtNextBibliography{\small}
\printbibliography}

\end{document}

%% file: Commands.tex
\newcommand{\ignore}[1]{}

\newcommand{\norm}[1]{\left\Vert#1\right\Vert} 
\newcommand{\mc}[1]{\mathcal{#1}}

\newcommand{\bma}[1]{\left[\begin{array}{ #1}}
\newcommand{\ema}{\end{array}\right]}

\DeclareMathAlphabet{\mbf}{OT1}{ptm}{b}{n}
\newcommand{\mbs}[1]{{\boldsymbol{#1}}}

\newcommand{\mbsdot}[1]{{\dot{\boldsymbol{#1}}}}

\newcommand{\mbscheck}[1]{{\check{\boldsymbol{#1}}}}

\newcommand{\mbfdot}[1]{{\dot{\mbf{#1}}}}
\newcommand{\mbfbar}[1]{{\bar{\mbf{#1}}}}
\newcommand{\mbfhat}[1]{{\hat{\mbf{#1}}}}
\newcommand{\mbfcheck}[1]{{\check{\mbf{#1}}}}

\def\fdota{{\raisebox{-2pt}{\LARGE $\cdot$}}}
\def\fdotb{{\raisebox{-0.6ex}{ \kern0.2ex\raisebox{0.8ex}{\tiny $\hspace*{-1ex}\circ$}}}}
\def\fddota{{\raisebox{-2pt}{\LARGE $\cdot\hspace*{-0.2ex}\cdot$}}}
\def\fddotb{{\raisebox{-0.6ex}{ \kern0.2ex\raisebox{0.8ex}{\tiny $\hspace*{-1ex}\circ\circ$}}}}

\newcommand{\dee}{\textrm{d}}

\newcommand{\trans}{{\ensuremath{\mathsf{T}}}} 
\newcommand{\utimes}{ {\raisebox{-0.6ex}{ \kern-1.0ex\raisebox{0.6ex}{ \small $\mathsf{v}$}}} } %
\newcommand{\beq}{\begin{equation}}
\newcommand{\eeq}{\end{equation}}
\newcommand{\bdis}{\begin{displaymath}}
\newcommand{\edis}{\end{displaymath}}
\newcommand{\beqarray}{\begin{eqnarray}}
\newcommand{\eeqarray}{\end{eqnarray}}
\newcommand{\beqarraynn}{\begin{eqnarray*}}
\newcommand{\eeqarraynn}{\end{eqnarray*}}
\newcommand{\balign}{\begin{align}}
\newcommand{\ealign}{\end{align}}
\newcommand{\balignnn}{\begin{align*}}
\newcommand{\ealignnn}{\end{align}}

\renewcommand{\theenumii}{\arabic{enumii}}
\makeatletter
\renewcommand{\p@enumii}{\theenumi.}
\makeatother


%% file: arxiv_cover_ieee.tex
%
%
%
%
%
%
%
\def \myJournal {IEEE/RSJ International Conference on Intelligent Robots and Systems (IROS)}
\def \myDoi {}
\def \myPaperSiteName {}
\def \myPaperSiteLink {}
\def \myYear {2021}
\def \myPaperCitation{C. C. Cossette, M. Shalaby, D. Saussié, and J. R. Forbes, ``Localization with Directional Coordinates (preprint),'' in \textit{2021 IEEE/RSJ International Conference on Intelligent Robots and Systems}, 2021.}


\begin{figure*}[t]

\thispagestyle{empty}
\begin{center}
\begin{minipage}{6in}
\centering
This paper has been accepted for publication in \emph{\myJournal}. 
\vspace{1em}

This is the author's version of an article that has, or will be, published in this journal or conference. Changes were, or will be, made to this version by the publisher prior to publication.
\vspace{2em}


\vspace{2em}
Please cite this paper as:

\myPaperCitation

\vspace{15cm}
\copyright \myYear \hspace{4pt}IEEE. Personal use of this material is permitted. Permission from IEEE must be obtained for all other uses, in any current or future media, including reprinting/republishing this material for advertising or promotional purposes, creating new collective works, for resale or redistribution to servers or lists, or reuse of any copyrighted component of this work in other works.

\end{minipage}
\end{center}
\end{figure*}
\newpage
\clearpage
\pagenumbering{arabic}